\documentclass[letterpaper]{article} 
\usepackage{aaai23}  
\usepackage{times}  
\usepackage{helvet}  
\usepackage{courier}  
\usepackage[hyphens]{url}  
\usepackage{graphicx} 
\usepackage{balance}
\usepackage{comment}

\usepackage{natbib}  
\usepackage{caption} 
\usepackage{amsmath,amssymb,amsfonts}
\usepackage{amsthm}
\usepackage{accents}
\usepackage{longtable,tabularx,booktabs}
\usepackage{makecell}
\usepackage{multirow}
\usepackage[flushleft]{threeparttable}
\usepackage{bbding}
\usepackage{subfigure,bigstrut}
\usepackage[linesnumbered,ruled,vlined]{algorithm2e}
\SetKwInput{KwInput}{Input}     
\SetKwInput{KwOutput}{Output}   

\newcommand{\ie}{{\em i.e.}}
\newcommand{\etal}{{\em et~al.}}
\newcommand{\ubar}[1]{\underaccent{\bar}{#1}}

\newtheorem{theorem}{Theorem}
\newtheorem{remark}{Remark}

\newtheorem{lemma}{Lemma}

\theoremstyle{definition}
\newtheorem{definition}{Definition}

\newtheorem{manualtheoreminner}{Lemma}
\newenvironment{manualtheorem}[1]{%
  \renewcommand\themanualtheoreminner{#1}%
  \manualtheoreminner
}{\endmanualtheoreminner}

\newtheorem{manualtheoreminnerr}{Theorem}
\newenvironment{manualtheorem1}[1]{%
  \renewcommand\themanualtheoreminnerr{#1}%
  \manualtheoreminnerr
}{\endmanualtheoreminnerr}

\newcommand{\pluseq}{\mathrel{+}=}

\title{Towards Verifying the Geometric Robustness of Large-Scale Neural Networks}
\author {
    Fu Wang\textsuperscript{\rm 1},
    Peipei Xu\textsuperscript{\rm 2},
    Wenjie Ruan\textsuperscript{\rm 1}\footnote{Corresponding Author},
    Xiaowei Huang\textsuperscript{\rm 2}
}
\affiliations {
    \textsuperscript{\rm 1} Department of Computer Science, University of Exeter, Exeter, EX4 4QF, UK\\
    \textsuperscript{\rm 2} Department of Computer Science, University of Liverpool, Liverpool, L69 3BX, UK\\
    \{fw377, w.ruan\}@exeter.ac.uk, \{peipei.xu, xiaowei.huang\}@liverpool.ac.uk\\
}

\begin{document}

\maketitle

\begin{abstract}

Deep neural networks (DNNs) are known to be vulnerable to adversarial geometric transformation.
This paper aims to verify the robustness of large-scale DNNs against the combination of multiple geometric transformations with a provable guarantee. 
Given a set of transformations (e.g., rotation, scaling, etc.), we develop GeoRobust, a black-box robustness analyser built upon a novel global optimisation strategy, for locating the {\em worst-case} combination of transformations that affect and even alter a network's output. 
GeoRobust can provide {\em provable guarantees} on finding the {\em worst-case} combination based on recent advances in Lipschitzian theory. 
Due to its black-box nature, GeoRobust can be deployed on large-scale DNNs regardless of their architectures, activation functions, and the number of neurons.
In practice, GeoRobust can locate the worst-case geometric transformation with high precision for the ResNet50 model on ImageNet in a few seconds on average.
We examined 18 ImageNet classifiers, including the ResNet family and vision transformers, and found a positive correlation between the geometric robustness of the networks and the parameter numbers. 
We also observe that increasing the depth of DNN is more beneficial than increasing its width in terms of improving its geometric robustness.
Our tool \textbf{GeoRobust} is available at \url{https://github.com/TrustAI/GeoRobust}.
\end{abstract}

\section{Introduction}

Although deep neural networks have achieved human-level performance, concerns are raised about their safety and reliability~\cite{HuangKR+18,ruan2019global,wang2022deep,ruan2021adversarial,wu2020game}.
In computer vision tasks, while deep learning models are known to be vulnerable to adversarial perturbations in pixel values~\cite{SzegedyZSBEGF13,croce2020reliable,yin2022dimba,mu2021sparse,mu20223dverifier},~\citet{EngstromTTSM19} show that a slight rotation of an input example can also fool DNNs.
Although modern DNNs are believed to be able to learn geometric information from training data~\cite{BakryEEE15}, they are not yet invariant to simple adversarial geometric transformations \cite{zhang2020generalizing}.

\begin{figure*}[!t]
  \centering
    \includegraphics[width=0.9\textwidth]{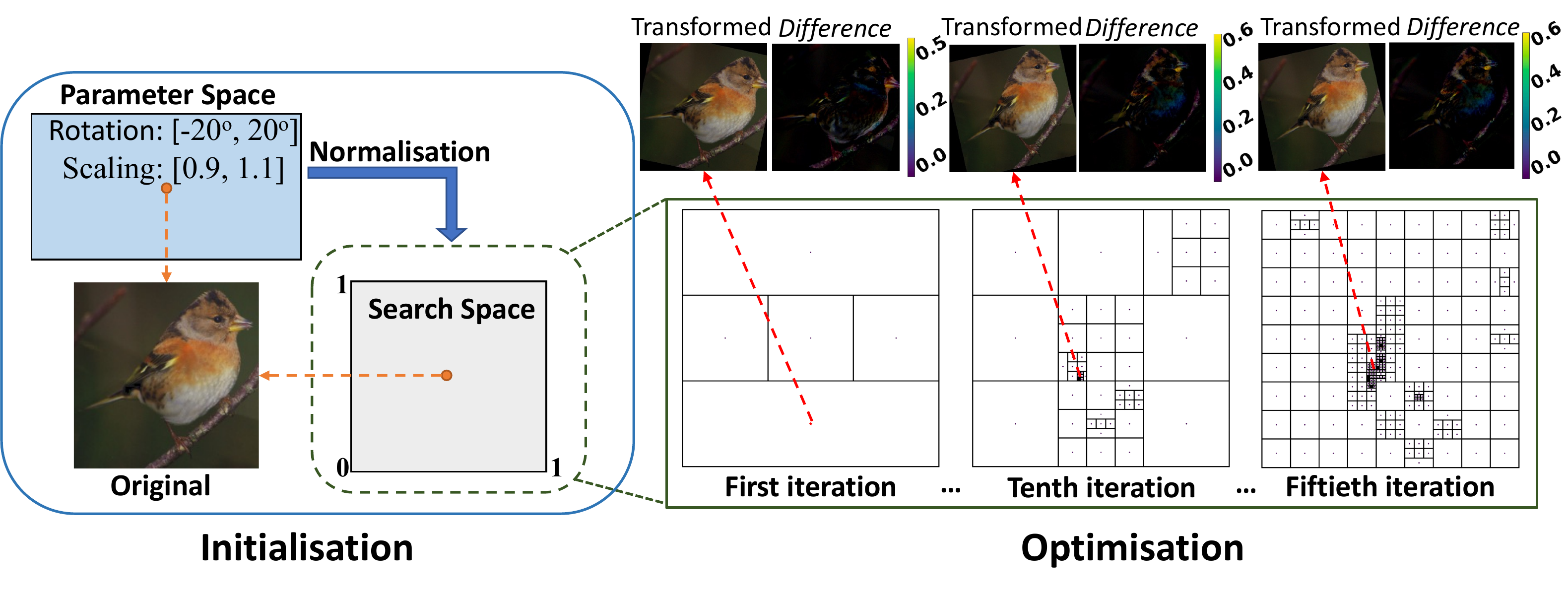}
    \vspace{-4mm}
    \caption{Schematic illustration of GeoRobust framework.
    After normalising the parameter space to a unit search space, GeoRobust performs a sequence of space divisions to find the global worst-case transformation.}
    \label{fig:pipeline}
    \vspace{-4mm}
\end{figure*}

Although additive adversarial perturbation has received tremendous attention, geometric transformations are more common and applicable in the physical world but have been less studied \cite{zhang2023generalizing}.
There is no efficient solution for searching the {\em worst-case} adversarial transformation with {\em provable guarantees} for {\em large-scale} DNNs.
\citet{EngstromTTSM19} showed that, although adversarial geometric transformation can be discovered through the random pick, it is a highly non-convex task that gradient-ascent-based adversarial attacks perform poorly.
\citet{PeiCYJ17} adopt the exhaustive search to find the worst-case transformation that alters the target model's prediction, but its computational complexity grows exponentially with the dimension of the considered transformations. 
Some researchers have adopted verification techniques~\cite{weng2018FastLin,SinghGPV19,CohenRK19} to analyse geometric transformations.
Relaxation-based approaches~\cite{weng2018FastLin,BalunovicBSGV19,MohapatraWCLD20} require the $L_p$-norm based constraint on pixel space for an input example.
However, as shown in Fig.~\ref{fig:pipeline}, geometric transformation can significantly change the values of the pixels, leading to severe violence against the constraint in the pixel space while still preserving human imperceptibility.
Therefore, the scalability of $L_p$ norm-based verification is limited for dealing with geometric transformation.
Recently, \citet{FischerBV20,LiWXRK00021} showed that randomised smoothing could be utilised to verify robustness against a single geometric transform, but their methods cannot handle the combination of multiple geometric transformations.
In addition, current methods can only provide a verifiable lower bound for verification purposes but cannot identify the {\em worst-case} geometric transformation that would actually minimise the model's confidence and potentially alter its prediction.

In this paper, we develop a novel black-box evaluation framework, GeoRobust, to study the geometric robustness,~\ie, the robustness of the model against adversarial geometric transformations.
GeoRobust takes advantage of both recent developments in Lipschitzian optimisation methods~\cite{Jones1993Lipschi,Gablonsky01} that provide provable guarantees on locating the worst-case transformation and the efficient parallel computation on Graphic Processing Units (GPUs).
The workflow of GeoRobust is presented in Fig.~\ref{fig:pipeline}. 
Given a set of geometric transformations and an input example,
GeoRobust converges to the worst-case combination for minimising an adversarial loss within a finite number of queries.
We enable GeoRobust to better utilise GPUs by easing its sampling condition.
Besides, a lower bound estimation method is also introduced to make GeoRobust an {\em anytime} verification, which can produce the lower and upper bound of the worst-case loss value whenever the algorithm stops.

In summary, our key contributions lie in three aspects.
\begin{enumerate}
    \item We prove the geometric transformation done by spatial transformation network (STN)~\cite{JaderbergSZK15} is Lipschitz continuous. By stacking STN module in front of Lipschitz-continuous neural networks, we can analyse their geometric robustness with guaranteed convergence.
    
    \item We develop GeoRobust, a black-box geometric robustness analyser, by taking advantage of Lipschitzian optimisation theory~\cite{Jones1993Lipschi}. 
    The convergence of GeoRobust is theoretically guaranteed, and it is also highly efficient in practice. 
    In our experiment, GeoRobust could find the worst-case adversarial transformations on an ImageNet image to evaluate a ResNet50 classifier with desirable precision in seconds.
    
    \item We use GeoRobust to benchmark the geometric robustness of state-of-the-art ImageNet classifiers, including the ResNet family and vision transformers. There are two main takeaways from our experiments: {\em i)} the geometric robustness of DNNs has a positive correlation with the number of parameters; and {\em ii)} increasing the number of layers seems to be more effective than adding more hidden units in each layer in improving the geometric robustness of DNNs.
\end{enumerate}

\section{Preliminaries}

\paragraph{Lipschitz continuity and Lipschitzian optimisation}
Previous studies indicate that the majority of modern DNNs are Lipschitz continuous~\cite{SzegedyZSBEGF13,RuanHK18,VirmauxS18,zhang2023model}.
The Lipschitz constant for a DNN gives an upper bound on how fast its output could change when small perturbations are applied to its input~\cite{SzegedyZSBEGF13,RuanHK18}.
Such a concept is closely related to the robustness of DNN, but exactly computing the smallest Lipschitz constant of a DNN is proven to be an NP-hard problem~\cite{VirmauxS18}.


Relying on the Lipschitz continuity, Lipschitzian optimisation is a query-based optimisation method that uses the Lipschitz constant of the objective function to gradually narrow the search space and locate the global optimum~\cite{piyavskii1972algorithm,shubert1972sequential,RuanHK18,xu2022quantifying,zhang2023reachability}.
While the Lipschitz constant of the objective function is necessary for classic Lipschitzian optimisation, a novel Lipschitzian optimisation solution, DIRECT~\cite{Jones1993Lipschi,jones2021direct}, does not require the Lipschitz constant to find the global optimum.
As detailed in the methodology section, we improve the DIRECT method for studying the geometric robustness of DNNs.

\paragraph{Geometric transformations}

Geometric transformations are element-wise manipulation that can be conducted via several physically meaningful parameters~\cite{szeliski2022computer}.
Given an image example, $x \in \mathbb{R}^{H\times W\times C}$ with height $H$, width $W$, and colour channels $C$, the geometric transformation $T_{\theta}$ is carried out on each channel equally.
Let $x_\mathtt{c} \in \mathbb{R}^{H\times W}$ be any channel of $x$ and the output of $T_{\theta}$ be $x^\prime_\mathtt{c}$.
For a pixel in $x^\prime_\mathtt{c}$ with index $(\mathtt{x}^\prime_i, \mathtt{y}^\prime_i)$, its value $V_i$ is mapped to the pixel indexed by $(\mathtt{x}_j, \mathtt{y}_j)$ in $x_\mathtt{c}$ via a transformation matrix $A_\theta$, \ie,
\begin{equation}\small
\left[\begin{array}{l}
\mathtt{x}_{j} \\
\mathtt{y}_{j}
\end{array}\right]
=A_{\theta}\left[\begin{array}{c}
\mathtt{x}_{i}^\prime \\
\mathtt{y}_{i}^\prime \\
1
\end{array}\right]=
\left[\begin{array}{lll}
\theta_{11} & \theta_{12} & \theta_{13} \\
\theta_{21} & \theta_{22} & \theta_{23}
\end{array}\right] \left[\begin{array}{c}
\mathtt{x}_{i}^\prime \\
\mathtt{y}_{i}^\prime \\
1
\end{array}\right].
\end{equation}
We adopt Spatial Transformation Network (STN)~\cite{JaderbergSZK15} to conduct the geometric transformation and use the bilinear sampling kernel to handle the projection of the non-integer index, which gives the transformation result:
\begin{equation}\small
\begin{aligned}
V_i&=\\
&\sum_{h}^{H} \sum_{w}^{W} U_{h w} \max \left(0,1-\left|\mathtt{x}_{i}-w\right|\right) \max \left(0,1-\left|\mathtt{y}_{i}-h\right|\right),
\end{aligned}
\end{equation}
where $U_{hw}$ denotes the value of a pixel, indexed by $(\mathtt{x}_h, \mathtt{y}_w)$, in $x_\mathtt{c}$, and the index does not need to be an integer.

\section{Problem Formulation}
\label{sec:4_formulation}

Given a neural network $F:\mathbb{R}^N\rightarrow \mathbb{R}^K$, an input example $x \in \mathbb{R}^N$, and its label $y \in \{1,\ldots, K\}$, we aim to find the optimal combination of several geometric transformations $T_{\theta}$ that can minimise $\ell:\mathbb{R}^n\rightarrow \mathbb{R}$,~\ie,
\begin{equation}\small
\label{eqn:goal}
\min\limits_{\theta \in\Theta} \;\ell(\theta;F, x, y),
\end{equation}
where $\Theta$ is the adversarial space that contains all feasible $\theta$, and $\ell$ denotes the margin loss defined as
\begin{equation}\small
\label{eqn:loss}
    \ell(\theta;F, x, y) = F_y(T_\theta(x)) - \underset{k \in \{1,\ldots, K\}\backslash \{y\}}{\max}\; F_k(T_\theta(x)),
\end{equation}
which allow us to determine whether the model $F$ can be fooled by $T_\theta(x)$ via verifying the lower bound of $\ell$. 
Specifically, if 
$
  \inf_{\theta \in \Theta} \ell(\theta;F, x, y) > 0 
$
is satisfied, the robustness of the model $F$ on the example $x$ would be verified.

Regarding geometric transformation, we consider rotation, translation, and isotropic scaling.
The corresponding transformation matrix $A_\theta$ can be written as
\begin{equation}\label{eqn:matrixA}\small
A_\theta 
= 
\left[\begin{array}{lll}
\theta_{11} & \theta_{12} & \theta_{13} \\
\theta_{21} & \theta_{22} & \theta_{23}
\end{array}\right] 
= 
\left[\begin{array}{lll}
\lambda\cos\gamma \;\;& -\sin\gamma \;\;& t^{hor} \\
\sin\gamma \;\;& \lambda\cos\gamma \;\;& t^{vrt}
\end{array}\right],
\end{equation}
where $\lambda$ is the scaling factor, $\gamma$ is the rotation angle, and $t^{hor}$ and $t^{vrt}$ control the horizontal and vertical translation.

\section{Methodology}

This section introduces a Lipschitzian optimisation-based approach, GeoRobust, to search for the worst-case transformation.
GeoRobust is composed of four components: 
\textit{1)} a Lipschitz continuous module for performing geometric transformations;
\textit{2)} a space division procedure; 
\textit{3)} a mechanism to select Potential Optimal (PO) subspaces that are more likely to contain the global minimum points than others; 
\textit{4)} and an anytime estimation of the global minimum.
While the space division procedure and the PO subspace selection are adopted from the DIRECT algorithm, we extend the definition of PO subspace and encourage the algorithm to query more subspaces at each iteration.
Because all evaluations at the same iteration can be done parallelly in a single forward propagation, our method could reach convergence within reduced iterations.
Furthermore, GeoRobust can estimate the model's Lipschitz constant and produce a reasonable lower bound of the given loss function. 
The pseudocode of GeoRobust and related proofs are provided in \textbf{Appendix}\footnote{Available at \url{https://github.com/TrustAI/GeoRobust.}}.

\paragraph{Notations}
Given a matrix $P \in \mathbb{R}^{2 \times n}$, one can define an $n$-dimensional parameter space, in which the upper and lower bounds on $i$-th transformation factors are given by $P_i$, where $i \in \{1,\ldots,n\}$.
GeoRobust normalises the parameter space into an $n$-dimensional unit hypercube, namely the search space, whose centre point corresponds to the identical transformation.
For a hyperrectangle with index $q$, denoted by $H_q$, in the search space, the value of the objective function at its centre point $c_q$ is denoted by $\ell(c_q)$.
We denote by $l^q_i$ the side length w.r.t. the $i$-th dimension, and by $\boldsymbol{e}_i$  the unit vector along $i$-th dimension.
The size of $H_q$ is defined as $\sigma_q = \max_{i \in \{1,\ldots,n\}} \frac{1}{2}l^q_i $, which is the same $L_\infty$ norm based measurement used by~\citet{Gablonsky01}.
For all hyperrectangles within the search space, we denote by $\mathcal{H}$ the set of hyperrectangles' indexes, and by $\ell_{\min} = \min_{p \in \mathcal{H}} \ell(c_p)$ the current best query result.
The Lipschitz constant of the objective function w.r.t. the search space is denoted by $\tilde{K}$.
GeoRobust computes the slope $\hat{K}$ between queried points w.r.t. the parameter space during optimisation and produces $\ell^{*}_{\min}$, an estimation of the lower bound of $\ell_{\min}$.

\subsection{Geometric Transformations Module}

The convergence guarantee of GeoRobust is related to the Lipschitz continuity of the target model. 
We give the following lemma to show that geometric transformations with bilinear sampling kernel are Lipschitz continuous,
which means, as long as a DNN model is Lipschitz continuous, stacking a geometric transformation module in front of it would not compromise the Lipschitz continuity~\cite{TsuzukuSS18}.
\begin{lemma}~\label{lem:lip_stn}
Given an input image example $x \in \mathbb{R}^{H\times W\times C}$ and the ranges of transformation factors, the first-order derivative of geometric transformation with bilinear sampling w.r.t. each transformation factor is bounded.
\end{lemma}

\subsection{Finding Optimal Geometric Transformation}

GeoRobust first divides the search space into subspaces according to the query results at their centre points. Then some subspaces that are more likely to contain the global minimum than others will be chosen as PO subspaces.
GeoRobust separates selected PO spaces and identifies new ones throughout each iteration of the optimisation process till the termination criteria are satisfied.

\paragraph{Space division}

As the only hypercube after initialisation, the initial PO subspace is the united search space itself.
GeoRobust {\bf trisects} the subspace and assigns the generated new subspace according to the query result, where the larger hyperrectangles include the better query result.
Without loss of generality, let $H_p$ be a PO subspace, which is an $n$-dimensional hyperrectangle containing $m$ dimensions with long sides of a length $3^{-d}$, where $m \leq n$, and $n - m$ dimensions with short sides of a length $3^{-d-1}$. 
GeoRobust ignores short sides and queries the value of the object function at the points $c\pm 3^{-d-1} \boldsymbol{e}_i$, where $i \in \{1,\ldots,m\}$.
For each dimension with long sides, the best query result is given by
\begin{equation}
    w_i = \min \big( \ell(c+3^{-d-1} \boldsymbol{e}_i), \ell(c- 3^{-d-1} \boldsymbol{e}_i) \big).
\end{equation}
As GeoRobust performs trisection only during division, the sizes of new subspaces are deterministic.
By dividing the above $H_p$, GeoRobust creates $2m+1$ new subspaces, including 3 sub-hypercubes with the side length of $3^{-d-1}$ and $m-1$ pairs of hyperrectangles, which have $1$ to $m-1$ dimensions with long sides of length $3^{-d}$.
The point corresponding to the best query result, $\min_{i \in \{1,\ldots,m\}} w_i$, is a centre point of a hyperrectangle with $m-1$ long sides.
This space division procedure is visualised in {\bf Appendix}.
Overall, such a division strategy encourages GeoRobust to divide the search space uniformly and further explore the area around the current best result, which we will detail later in the PO subspace selection.

\paragraph{Identifying potential optimal subspaces}
The space division procedure creates new subspaces, and the next step is to locate new PO subspaces for further division.
Ideally, a PO hyperrectangle $H_p$ is expected to satisfy two conditions~\cite{Jones1993Lipschi}:
\begin{align}
    &\ell\left(c_{p}\right)-\tilde{K} \sigma_{p} \leq \ell\left(c_{q}\right)-\tilde{K} \sigma_{q}, \forall q \in \mathcal{H}, \label{eqn:po_cond1}\\
&\ell\left(c_{p}\right)-\tilde{K} \sigma_{p} \leq \ell_{\min }-\tau\left|\ell_{\min }\right|.\label{eqn:po_cond2}
\end{align}
Inequation~\eqref{eqn:po_cond1} indicates that only the hyperrectangles with the potential to improve the current $\ell_{\min}$ can be chosen as PO subspaces.
Meanwhile, the second condition~\eqref{eqn:po_cond2} ensures that the possible improvement in the chosen subspaces is greater than $\tau\left|\ell_{\min }\right|$, where a reasonable choice of $\tau$ is between $10^{-3}$ and $10^{-7}$~\cite{JonesM21}.
Taking advantage of DIRECT optimisation, GeoRobust does not need to know the Lipschitz constant. The following lemma demonstrates how to search for PO hyperrectangles in the absence of $\tilde{K}$.

\begin{lemma}\cite{Gablonsky01}~\label{lem:find_po}
Given the index set $\mathcal{H}$ and a positive tolerance $\tau > 0$. Let $\ell_{\min}$ denote the current best query result. 
Let $\mathcal{H}^p_1 = \{q\in \mathcal{H}:\sigma_q < \sigma_p\}$, $\mathcal{H}^p_2 = \{q\in \mathcal{H}:\sigma_q > \sigma_p\}$ and $\mathcal{H}^p_3 = \{q\in \mathcal{H}:\sigma_q = \sigma_p\}$.
A hyperrectangle $H_p$ is said to be potentially optimal if 
\begin{equation}\label{eqn:po_cond1_h3}
    \ell(c_p) \leq \ell(c_q), \forall q \in \mathcal{H}^p_3,
\end{equation}
and there is a $\tilde{K} > 0$ such that
\begin{equation}\small\label{eqn:po_cond1_without_lip}
\max _{q \in \mathcal{H}^p_{1}} \frac{\ell\left(c_{p}\right)-\ell\left(c_{q}\right)}{\sigma_{p}-\sigma_{q}} \leq \tilde{K} \leq \min _{q \in \mathcal{H}^p_{2}} \frac{\ell\left(c_{q}\right)-\ell\left(c_{p}\right)}{\sigma_{q}-\sigma_{p}},
\end{equation}
and 
\begin{equation}\small\label{eqn:po_cond2_}
 \begin{cases}
    \tau \leq \frac{\ell_{\min }-\ell\left(c_{p}\right)}{\left|\ell_{\min }\right|}+\frac{\sigma_{p}}{\left|\ell_{\min }\right|} \min _{q \in \mathcal{H}^p_{2}} \frac{\ell\left(c_{q}\right)-\ell\left(c_{p}\right)}{\sigma_{q}-\sigma_{p}},   &  \text{if $\ell_{\min } \neq 0$,} \\
    \ell\left(c_{p}\right) \leq \sigma_{p} \min _{q \in \mathcal{H}^p_{2}} \frac{\ell\left(c_{q}\right)-\ell\left(c_{p}\right)}{\sigma_{q}-\sigma_{p}},
        &  \mbox{otherwise.}
 \end{cases}                
 \end{equation}
\end{lemma}

\subsection{Generalising PO Conditions to Unleash the Power of Parallel Computation}
The conditions in Lemma~\ref{lem:find_po} are meant to select a small number of subspaces to reduce the total number of function evaluations, which is typically the most time-consuming procedure.
However, by leveraging modern deep learning frameworks, one can easily query multiple examples on a target DNN via a single forward propagation on GPUs, where the computational time difference between evaluating a single sample and a batch of samples is marginal.
Therefore, relaxing the constraint given by inequation~\eqref{eqn:po_cond1_h3}, we define the following $\alpha$ candidate set to select more subspaces.

\begin{definition}[$\alpha$ candidate set]\label{def:alpha_set}

Following the same notions in Lemma~\ref{lem:find_po}, we define the $\alpha$ candidate set as
\begin{equation}\small
    \mathcal{H}_\alpha=
     \begin{cases} 
     \varnothing,&\mbox{if} \max_{p \in\mathcal{H}^p_3} s_p \leq 0,\\
     \{p_1,\ldots,p_{\alpha'}: \max\sum_{j=1}^{\alpha'}s_{p_j}\},&\mbox{otherwise,}
     \end{cases}
\end{equation}
where $\alpha' \leq \alpha$ and the optimal score $s_{p}$ is given by
\begin{equation}\small\label{eqn:po_score}
    s_{p} = \min _{q \in \mathcal{H}^p_{2}} \frac{\ell\left(c_{q}\right)-\ell\left(c_{p}\right)}{\sigma_{q}-\sigma_{p}}
    -
    \max _{q \in \mathcal{H}^p_{1}} \frac{\ell\left(c_{p}\right)-\ell\left(c_{q}\right)}{\sigma_{p}-\sigma_{q}}.
\end{equation}
\end{definition}

Modifying the condition~\eqref{eqn:po_cond1_without_lip} into the score described in Eq.~\eqref{eqn:po_score} allows us to rank the potential minimum contained by subspaces with the same size.
The proposed $\alpha$ candidate set is easy to control.
When $\alpha=1$, Definition~\ref{def:alpha_set} degrades to condition~\eqref{eqn:po_cond1_without_lip}, while increasing $\alpha$, GeoRobust explores more subspaces in each iteration.
As described in the next section, all subspaces will be subdivided by GeoRobust after a certain number of iterations.
Instead of only partitioning spaces satisfying inequation~\eqref{eqn:po_cond1}, $\alpha$ candidate set also selects hyperrectangles that would likely satisfy Lemma~\ref{lem:find_po} in the next few rounds.
While the number of queries would rise, using the $\alpha$ candidate set enables GeoRobust to discover the optimal subspace quickly.
On the other hand, because the function evaluations can be done in parallel on GPUs, replacing condition~\eqref{eqn:po_cond1_without_lip} with $\alpha$ candidate set only has a small influence on the computational time cost.

\subsection{Stop Criteria and Convergence Analysis}~\label{sec:43_convergence_stop}
In practice, GeoRobust is limited by three factors: the maximal number of iteration $T$,
the maximal number of queries $Q$, and the maximal number of trisection along each dimension, which is denoted by depth $D$.
The first two stop criteria are straightforward. 
We stop the optimisation once the computational budget runs out.
The limitation on depth is applied for two reasons~\cite{Gablonsky01}.
On the one hand, it puts a limitation on the smallest size of subspaces.
By doing so, GeoRobust is compelled to halt local search and encouraged to conduct global exploration when the current optimal subspace is sufficiently small, which accelerates the convergence.
On the other hand, defining the smallest subspace size also sets an upper bound for the number of queries.
For an $n$-dimensional search space, GeoRobust could conduct up to $3^{nD}$ times of queries, which is equivalent to a grid search. 
Besides, our implementation adopted the $L_\infty$ norm to measure hyperrectangles' size~\cite{Gablonsky01}. 
Such a measurement simplifies both space division and PO space selection. 
For the former, the $L_\infty$ norm is more computationally efficient than the Euclidean norm~\cite{JonesM21}.
For the latter, hyperrectangles are grouped by their longest side length under the $L_\infty$ norm, which introduces less number of different sizes to consider. 

\paragraph{Convergence analysis}

GeoRobust is guaranteed to converge to the global minimum within a pre-defined small tolerance after a finite number of queries if the objective function is continuous~\cite{Jones1993Lipschi}.
This guarantee comes from the following observation shown in Remark~\ref{rmk:all_divided}.

\begin{remark}\cite{Gablonsky01}\label{rmk:all_divided}
Following the same notions in Lemma~\ref{lem:find_po}, there is at least one hyperrectangle $H_p$ will be identified as PO subspace at each iteration, where $H_p$ satisfies $\mathcal{H}^p_2 = \varnothing$ and makes inequation~\eqref{eqn:po_cond1_h3} holds so that every hyperrectangle will be subdivided after finite iterations.
\end{remark}

Note that theoretically proving a specific DNN is Lipschitz continuous is beyond the scope of this paper, 
and existing works demonstrate the Lipschitz continuity of convolutional neural networks~\cite{RuanHK18} and vision transformers~\cite{VuckovicBT21,WangR22,wang2023ode4vitrobustness} used in the image classification task.
In addition, given by Lemma~\ref{lem:lip_stn}, we prove that the geometric transformation is Lipschitz continuous.
So, as long as the neural network satisfies a Lipschitz condition, the objective function~\eqref{eqn:goal} is Lipschitz continuous, and GeoRobust is guaranteed to locate to the global minimum after a sufficient number of queries.
The convergence complexity is described in Theorem~\ref{thm:convergence_complexity}.

\begin{theorem}\label{thm:convergence_complexity}
Let $C$ be the $n$-dimensional united search space and $\tilde{K}$ be the Lipschitz constant of $\ell$ w.r.t. $C$.
The gap between current minima and global minima after $T$ iterations can be written as
\begin{equation}\small\label{eqn:thm-1}
    \ell_{\min} - \min _{c \in C} \ell(c) \leq \varepsilon < \tilde{K}\cdot(T+1)^{-\frac{1}{n}}.
\end{equation}
Therefore, to achieve any desired $\varepsilon$, we need up to $\mathcal{O}\big((\tilde{K}/\varepsilon)^n\big)$ iterations.
\end{theorem}

\subsection{Estimating the Global Minimum}\label{sec:44_est}
While GeoRobust is guaranteed to find the global minimum eventually, we enable it to be an anytime analyser that can utilise intermediate query results to estimate the lower bound of the global minimum at each iteration.

Recall that the parameter space is defined via the matrix $P \in \mathbb{R}^{2\times n}$ that contains the upper and lower bounds of $n$ transformation factors.
To divide $m$ dimensions of a hyperrectangle, GeoRobust samples and evaluates new points at $c\pm 3^{-1} \cdot l_i \boldsymbol{e}_i$, where $i\in\{1,\ldots,m\}$.
The slopes along the $i$-th dimension are given by
\begin{equation*}
    \hat{K}_{c_i^+} = \frac{\|\ell(c) - \ell(c_i^+)\|}{d^c_i}\quad\mbox{and}\quad \hat{K}_{c_i^-} = \frac{\|\ell(c) - \ell(c_i^-)\|}{d^c_i},
\end{equation*}
where $c_i^+$ and $c_i^-$ are short hands for $c\pm 3^{-1} \cdot l_i \boldsymbol{e}_i$, and we denote by
$d^c_i = 3^{-1}\cdot l_i P_i\cdot\left[\begin{array}{c}1\\-1 \end{array}\right]$ the distance between $c$ and $c_i^{+/-}$ within the parameter space.
Then $\hat{K}_c$, the local slope of $c$, is updated to be the largest local slope,~\ie, $$\hat{K}_c = \max\{\hat{K}_{c_1^+},\hat{K}_{c_1^-},\ldots,\hat{K}_{c_m^+},\hat{K}_{c_m^-}\}.$$
GeoRobust updates the local slope after space division so that it can estimate the global minimum based on the query results at that time.
Let $c_o$ denote the centre of the current optimal hyperrectangle $H_o$ that has $\ell(c_o) = \ell_{\min}$ and $\hat{K} = \max_{q\in\mathcal{H}} \hat{K}_{c_q}$, the lower bound of global minimum can be estimated via
\begin{equation}\label{eqn:est_minimum}
    \ell^{*}_{\min} = \ell(c_o) - \hat{K}_{\max}\bar{\sigma}_o,
\end{equation}
where we take a relaxation on $\sigma_o$ and compute it via the Manhattan distance,~\ie, $\bar{\sigma}_o = \frac{1}{2}\sum_{i=1}^n d^o_i$.
Therefore, as long as GeoRobust can locate a $H_o$ containing the global minimum $\hat{\ell}_{\min}$, we have $\ell^{*}_{\min} \leq \hat{\ell}_{\min}$.

\section{Experiments}

Our experiments include three parts.
First, we compare GeoRobust to state-of-the-art baseline methods for verifying robustness against three geometric transformations,~\ie, rotation, translation, and scaling.
Then, we take advantage of GeoRobust to efficiently benchmark popular large-scale networks on ImageNet regarding their robustness over the combination of all three transformations.
Finally, we conduct an empirical analysis to study the impact of the depth and $\alpha$ conditions on the convergence of GeoRobust.

\begin{table*}[!t]\small
  \centering
    \begin{tabular}{c@{\hspace{2mm}}c@{\hspace{2mm}}c@{\hspace{2mm}}c@{\hspace{2mm}}c@{\hspace{2mm}}c@{\hspace{2mm}}c@{\hspace{2mm}}c@{\hspace{2mm}}c|c@{\hspace{2mm}}c}
    \toprule
    Dataset & Trans. & GeoRobust &Gsmooth & TSS &DeepG & Interval & Semantify-NN & DistSPT & TSS attack & Grid Search \\
    \midrule
    \multirow{2}[2]{*}{MNIST} & $R(50^{\circ})$ & \textbf{98.2\%}  & 95.7\% & 97.4\% & $\leq$ 85.8\%& $\leq$6.0\% & $\leq$ 92.48\% & 82\%  & 98.2\% & 98.2\%\\
           & $S(0.3)$ & \textbf{99.2\%}     &   95.9\% & 97.2\% & 85.0\% & 16.4\% & -     & -     & 99.2\% & 99.2\% \\
    \midrule
    \multirow{3}[2]{*}{CIFAR10} & $R(10^{\circ})$ & \textbf{74.8\%}     & 65.6\% & 70.6\% &  62.5\% & 20.2\% & -     & 37\%  & 76.4\% & 74.8\%  \\
          & $R(30^{\circ})$ & \textbf{66.4\%}     & -     & 63.6\% & 10.6\% & 0.0\%     & 49.37\% & 22\%  & 69.4\% & 66.4\%  \\
          &   $S(0.3)$   & \textbf{63.4\%}     &   54.3\%    & 58.8\% & 0.0\% & 0.0\%     & -     & -     & 67.0\% &63.4\% \\
    \bottomrule
    \end{tabular}%
    \caption{
  Comparing with baseline methods on MNIST and CIFAR-10 against rotation and Scaling. 
  We denote by $-$ an unsupported setting and by $0\%$ a failed verification.
  Baselines' performance is adopted from~\cite{LiWXRK00021,HaoYD0SZ22}.
  }\label{tab:baseline}
     \vspace{-2mm}
\end{table*}%

\paragraph{General setup} 
For geometric transformations, we denoted by $R(\gamma)$ the rotation angle between $\pm \gamma$, by $S(\lambda)$ the scaling range between $1\pm\lambda$, and by $T(t^{hor},t^{vrt})$ the translation that moves an example up to $t^{hor}$ and $t^{vrt}$ pixels horizontally and vertically, respectively. 
For the model architectures, the MNIST classifier is a network with four convolutional layers and three linear layers, and the CIFAR10 classifier's architecture is ResNet101.
We adopted the pre-trained MNIST and CIFAR10 models released by~\citet{LiWXRK00021} and utilised TIMM, a model zoo of ImageNet classifiers, to investigate the geometric robustness of popular large-scale DNNs.
Our experiment is performed on a machine with an Intel i7-10700KF CPU, an RTX 3090 GPU, and 48 gigabytes of memory.
More implementation details can be found in \textbf{Appendix}.

\subsection{Comparison with Previous Works}\label{sec:5_compare_imagenet}

Since GeoRobust only requires queries on the target models' output, it can be easily deployed on any pretrained neural network.
We follow the same setup used in TSS~\cite{LiWXRK00021} and GSmooth~\cite{HaoYD0SZ22} and apply GeoRobust on their pretrained models to conduct robustness verification on MNIST and CIFAR10. For GeoRobust, the robustness of an example is verified if the corresponding lower bound $\ell^*_{\min}>0$.
Because exhaustive search is computationally infeasible, we implement a grid search with a sufficient computational budget to run through the parameter space to test whether the model's prediction on each input example can be altered, which serves as the ground truth on the verified accuracy.
Please note that GeoRobust works on $L_\infty$-norm based parameter space, while~\citet{LiWXRK00021} uses $L_2$-norm based constraint on the translation, which is currently inapplicable for GeoRobust.
Therefore, the comparison is done on rotation and scaling transformations.
Although DeepG~\cite{BalunovicBSGV19} and TSS~\cite{LiWXRK00021} can analyse the geometric robustness of ImageNet classifiers, they are time-consuming and inefficient when dealing with transformation combinations.
Besides, we failed to properly reload the ImageNet classifiers evaluated by TSS (see {\bf Appendix} for details).
Therefore, the evaluation on ImageNet is done on a ResNet50 model, the same architecture used by~\citet{LiWXRK00021}, against different combinations of transformations, and we compare the performance with grid search and random pick.

\begin{figure}[!t]
    \centering
  \includegraphics[width=0.43\textwidth]{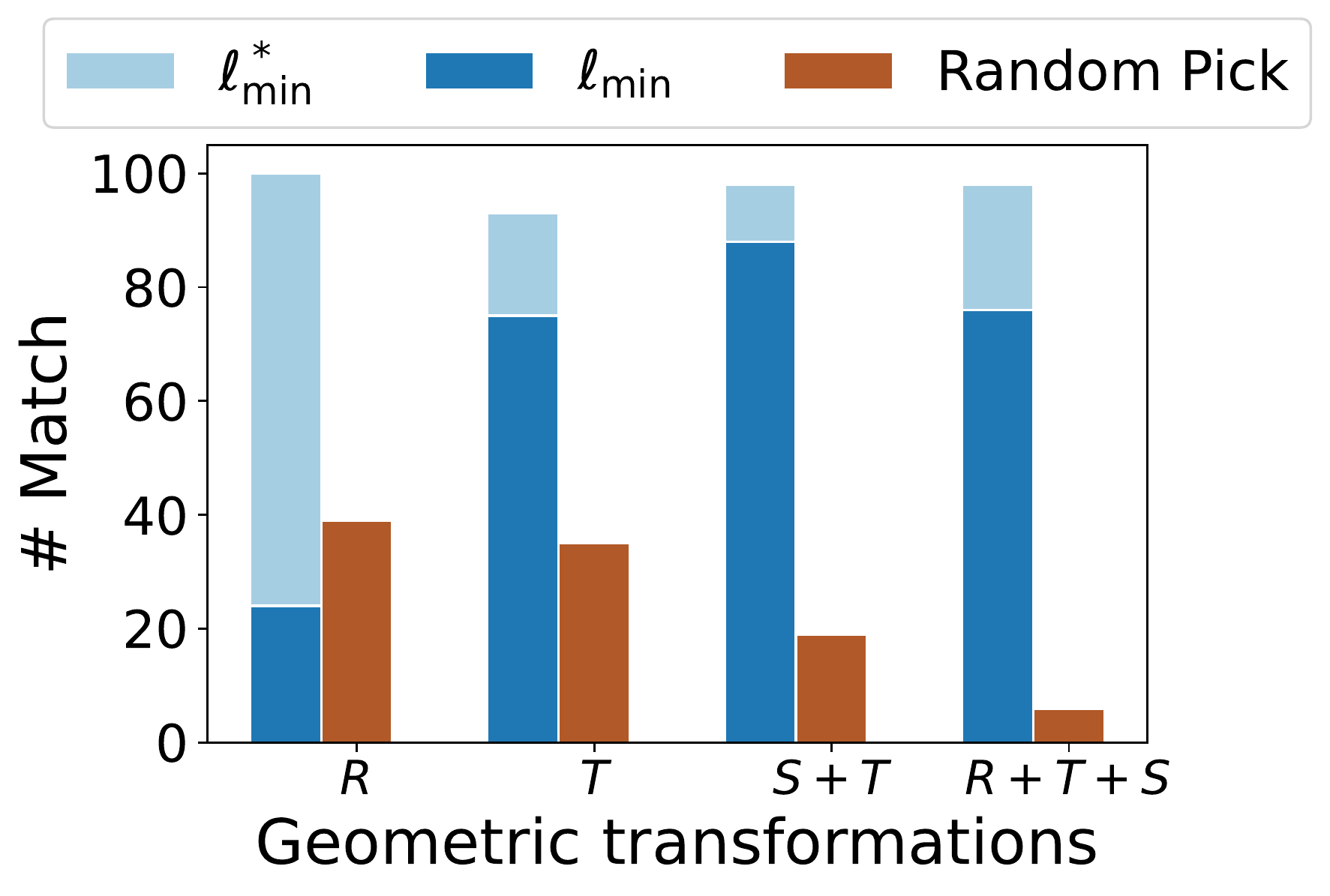}%
    \caption{Comparing the global minimum found by grid search, random pick, and GeoRobust. The geometric transformations are the same as in Tab.~\ref{tab:imagenet_baseline}.
    We mark an example as a match if its corresponding minimum found by a method is equal to or smaller than the minimum found by grid search.}
    \label{fig:resnet_match}
         \vspace{-2mm}
\end{figure}

The comparison results on MNIST and CIFAR10 are summarised in Tab.~\ref{tab:baseline}, in which we report the verified accuracy determined by each baseline method.
GeoRobust outperforms previous methods under all scenarios and reports the same verified accuracy as grid search.
The experiment demonstrates the effectiveness of GeoRobust in verifying the geometric robustness against 1-dimensional transformation.
The evaluation on ImageNet is summarised in Tab.~\ref{tab:imagenet_baseline}, in which we can see that the accuracy verified by GeoRobust is comparable to or better than the grid search. 
Random pick with sufficient queries can achieve a similar performance as grid search. Still, it tends to perform worse as the dimension of parameter space becomes larger.
In addition, as the minimum found by grid search is more likely to be the ground truth minimum, we mark an example as a match if its corresponding minimum found by a method is equal to or smaller than the minimum found by grid search.
As shown in Fig.~\ref{fig:resnet_match}, we can see that the estimated lower bound achieves considerable precision with a limited number of queries.
The performance of only verifying the translation is slightly worse than other transformations, where the reason might be the distortion introduced by bilinear sampling.

\begin{table}[!t]\small
    \centering
\begin{tabular}{ccccc}
    \toprule
    \multirow{2}[4]{*}{Methods} & \multicolumn{4}{c}{Transformation} \\
\cmidrule{2-5}    
    & $R$ & $T$ & $S+T$& $R+T+S$\\
    \midrule
    GeoRobust & 58\%    & 57\%     & 57\%     & 46\%  \\
    Random pick & 58\%     & 59\%     & 60\%     & 49\%  \\
    Grid search & 58\%    & 59\%     & 57\%     & 46\%  \\
    \bottomrule
    \end{tabular}%
    \caption{Verify geometric robustness on ImageNet with ResNet50 model, whose vanilla accuracy is 74\%, against $R(20^{\circ})$, $S(0.1)$, and $T(22.4,22.4)$ transformations.}\label{tab:imagenet_baseline}
     \vspace{-4mm}
\end{table}

\begin{figure*}[!t]
    \centering
    \subfigure[]
    {\label{fig:analysis_minimum}
    \includegraphics[width=0.66\textwidth]{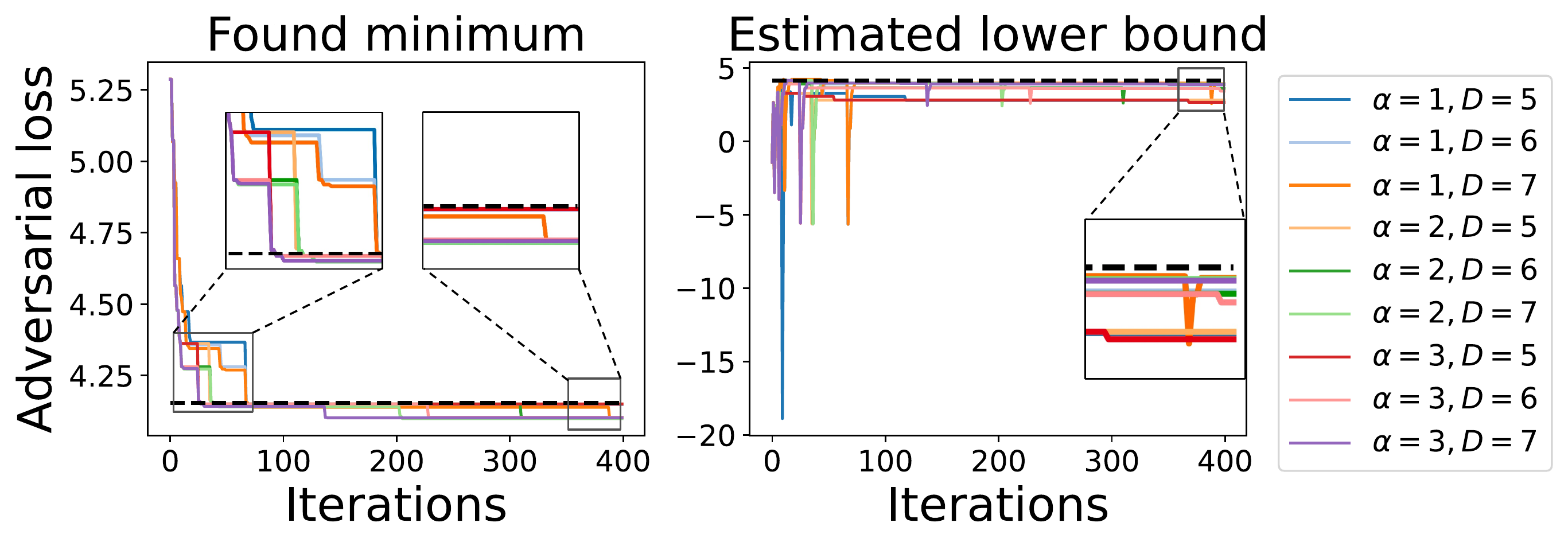}
    }
    \subfigure[]
    {\label{fig:analysis_cost}
    \includegraphics[width=0.31\textwidth]{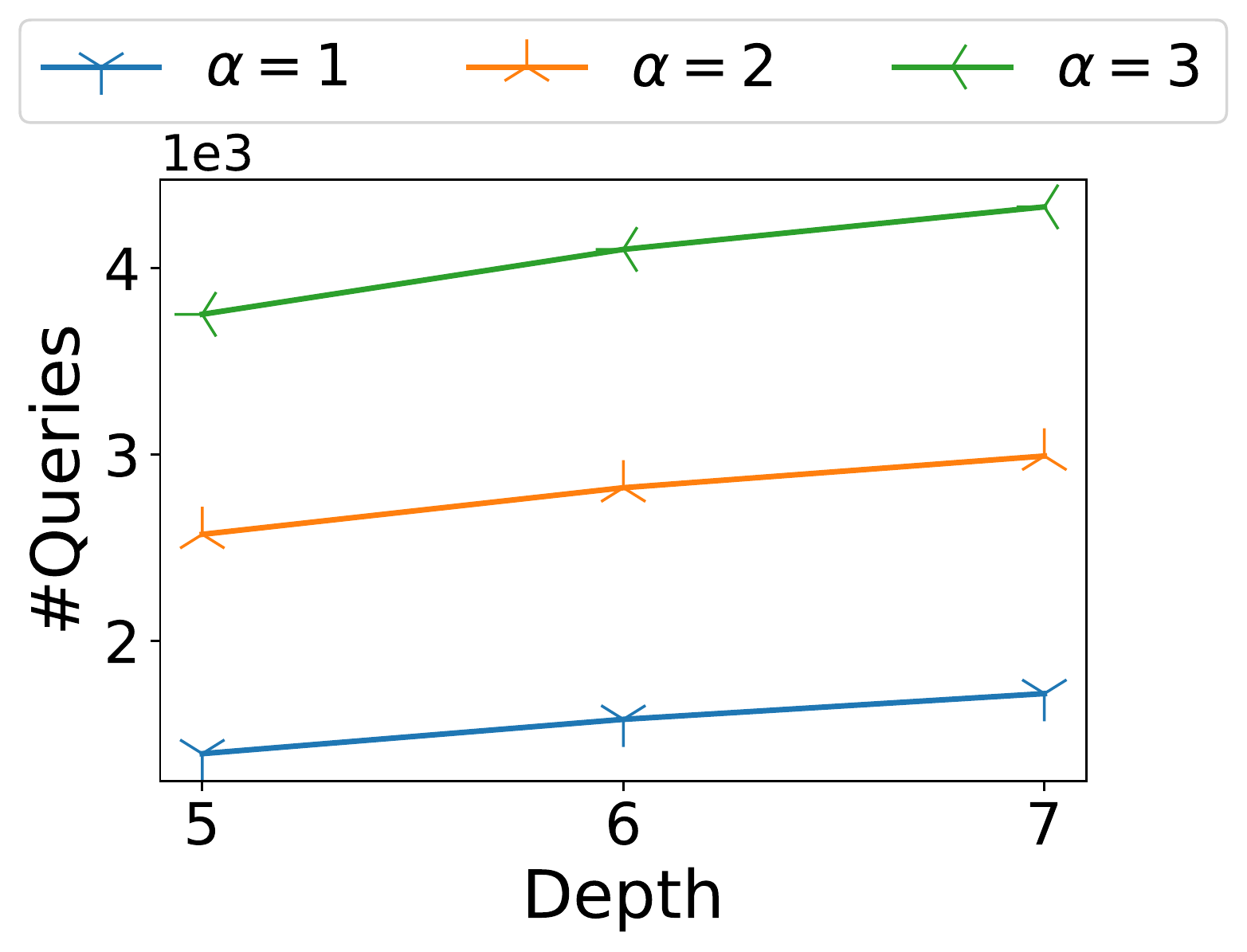}
    }
    \vspace{-4mm}
    \caption{
    Carrying out GeoRobust with different combinations of candidates set size~$\alpha$ and depth~$D$ on ResNet50.
    The black dot line in~\ref{fig:analysis_minimum} corresponds to a global minimum found in a grid search with $2.5\times10^5$ function evaluations.
    }\label{fig:analysis}
       \vspace{-4mm}
\end{figure*}

\paragraph{Runtime}
The effectiveness of GeoRobust is highly related to the transformation's dimensions.
In Tab.~\ref{tab:baseline}, GeoRobust is only performed on 1-dimensional transformation.
Its average runtime is 0.18 seconds and 0.72 seconds per example on MNIST and CIFAR10, respectively.
Furthermore, the average runtime for analysing the ResNet50 ImageNet classifier from 1-dimensional to 4-dimensional transformations are 2.4 seconds, 3.6 seconds, 4.0 seconds, and 4.5 seconds.
In comparison, according to~\cite{LiWXRK00021}, it takes TSS 17.7 and 1201.2 seconds, respectively, to analyse an MNIST example and an ImageNet example on the same model architectures w.r.t. a 1-dimensional transformation.

\begin{table}[!t]\small
\centering
\begin{tabular}{l@{\hspace{2mm}}c@{\hspace{2mm}}c@{\hspace{2mm}}c@{\hspace{2mm}}c}
\toprule[0.8pt]
Models & Vanilla& Attack & Verified& \#Parameters\\
\midrule
Inception v3.                    &73.60\%&28.20\%&24.20\%&2.4$\times10^7$\\
Inception v$3_{adv}$             &75.00\%&30.60\%&27.00\%&2.4$\times10^7$\\
Inception v4                     &78.40\%&40.20\%&36.40\%&4.3$\times10^7$\\
\midrule             
ResNet34                         &64.40\%&10.60\%&9.00\%&2.2$\times10^7$\\
ResNet50                         &78.40\%&54.00\%&31.12\%&2.6$\times10^7$\\
WideResNet50                 &81.60\%&49.40\%&40.00\%&6.9$\times10^7$\\
ResNet101                        &80.00\%&54.20\%&48.20\%&4.5$\times10^7$\\
ResNet152                        &79.40\%&53.80\%&46.20\%&6.0$\times10^7$\\
\midrule
$\mbox{Vit}_{32}$        &75.60\%&23.40\%&19.00\%&8.8$\times10^7$\\
$\mbox{Vit}_{16}$        &81.40\%&41.20\%&34.20\%&8.6$\times10^7$\\
Large $\mbox{Vit}_{16}$  &83.40\%&49.20\%&40.20\%&3.0$\times10^8$\\
\midrule
$\mbox{Beit}_{16}$.      &83.80\%&56.40\%&52.00\%&6.5$\times10^7$\\
Large $\mbox{Beit}_{16}$ &\textbf{85.60\%}&\textbf{65.60\%}&\textbf{58.20\%}&2.3$\times10^8$\\
\midrule
Gmlp                             &77.96\%&40.80\%&36.80\%&1.9$\times10^7$\\
Mixer                            &72.20\%&27.20\%&23.40\%&6.0$\times10^7$\\
Swin                             &80.20\%&34.60\%&13.20\%&8.8$\times10^7$\\
Xcit                             &76.80\%&40.40\%&20.60\%&8.4$\times10^7$\\
Pit                              &79.40\%&36.60\%&20.00\%&7.4$\times10^7$\\
\bottomrule
\end{tabular}
\caption{Benchmarking the geometric robustness of eighteen ImageNet classifiers.}\label{tab:benchmark_imagenet}
\vspace{-4mm}
\end{table}

\subsection{Benchmarking Geometric Robustness}

This section investigates the robustness of large-scale DNN classifiers against geometric transformations.
Since GeoRobust can efficiently locate a worst-case combination of transformations in a black-box manner, we utilise GeoRobust to evaluate the geometric robustness of large-scale ImageNet classifiers against the combination of rotation $R(20^\circ)$, translation $T(22.4,22.4)$, and scaling $S(0.1)$.

From Tab.~\ref{tab:benchmark_imagenet}, we can see that \textit{1)} models with more parameters appear to have better geometric robustness than those with fewer;
\textit{2)} widening a network seems less beneficial than deepening it in terms of improving the geometric robustness;
\textit{3)} the large version of Beit showed the best geometric robustness, whereas the basic Beit model is the second most robust model.
This phenomenon suggests that bidirectional modelling could be helpful for DNNs learning geometric information and obtaining geometric robustness;
\textit{4)} comparing the performance between Inception V3 and Inception V3$_{adv}$, an adversarially trained model, we can see that adversarial training does not significantly improve the model's geometric robustness.

\subsection{Empirical Analysis}
In Fig.~\ref{fig:analysis}, we carry out GeoRobust with different combinations of candidates set size~$\alpha$ and depth~$D$ on ResNet50.
Increasing the size of $\alpha$ candidates set enables GeoRobust to be more efficient in exploring the search space and locating the optimal subspace.
Due to the limitation on the subspaces' minimal size, as the depth gets larger, the optimal transformation combination found by GeoRobust is closer to the ground truth worst-case, and the estimated lower bound is closer to the global minimum as well. 
It can be observed that the upper bound remains unchanged after convergence, while the estimated lower bound would be updated whenever GeoRobust finds a larger local slope, which is why the estimations change in the right side plot of Fig.~\ref{fig:analysis_minimum}.
As shown in Fig.~\ref{fig:analysis_cost}, while the impact of depth on computational cost is trivial, increasing the $\alpha$ candidates set would significantly raise the total number of function queries in fixed iterations.
The runtime of GeoRobust with $\alpha=1$ and $D=5$ is 6.9 seconds, and the runtime is 16.1 seconds when it is carried out at $\alpha=3$ and $D=7$.
We can see that the runtime increases sub-linearly with the number of queries because the queries are done parallelly on GPUs.

\section{Related Works}
In this paper, we compared GeoRobust to Interval~\cite{SinghGPV19}, 
DeepG~\cite{BalunovicBSGV19}, 
Semantify-NN~\cite{MohapatraWCLD20}, 
TSS~\cite{LiWXRK00021}, 
GSmooth~\cite{HaoYD0SZ22}, 
and DistSPT~\cite{FischerBV20}.
DeepG~\cite{BalunovicBSGV19}, Semantify-NN~\cite{MohapatraWCLD20}, and Interval extend verification techniques designed for $L_p$-norm based additive perturbation.
Both Semantify-NN~\cite{MohapatraWCLD20} and GSmooth~\cite{HaoYD0SZ22} introduce small networks to simulate the geometric manipulation, where Semantify-NN adopts a linear relaxation-based verification~\cite{weng2018FastLin} and GSmooth applies random smoothing.
DistSPT~\cite{FischerBV20} and TSS~\cite{LiWXRK00021} are also randomised smoothing based approaches, where TSS is a black-box analyser that is scalable to large DNNs.
Besides, although the parameter space of control factors for most geometric manipulations is continuous, the image pixels' coordinates are bounded integers, which means the possible outcomes for a particular set of transformations are finite. 
Pei~\etal~\cite{PeiCYJ17} empirically evaluated the robustness of DNNs against geometric transformations by enumerating all possible values.
In contrast, our GeoRobust is a query-based black-box analyser that is fundamentally different to the above methods.
We demonstrated that as long as the target model is Lipschitz continuous, GeoRobust can verify the robustness of large-scale DNNs against a combination of geometric transformations in seconds.
The combination with probabilistic approaches~\cite{ZhangRF20} will be explored in our future works. 

\section{Conclusion}
In this paper, we propose a black-box analyser, GeoRobust, to efficiently verify the robustness of large-scale DNNs against geometric transformation.
Given the ranges of multiple geometric transformations, GeoRobust can find the worst-case manipulation that can minimise an adversarial loss without knowing the internal structures of the target model.
Theoretically, we prove the Lipschitz continuity of geometric transformations operated by STN and analyse the convergence complexity of GeoRobust.
On the methodology side, we generalise the sampling strategy to better leverage GPU parallel computation and design an anytime estimation method to approximate the lower bound.
With GeoRobust, we systematically benchmark the geometric robustness of popular ImageNet classifiers.
Our empirical study shows that larger neural networks are more robust against geometric manipulation.
Deepening a network improves its geometric robustness better than increasing its width.

\section*{Acknowledgements}
This work is supported by Partnership Resource Fund of ORCA Hub via the UK EPSRC under project [EP/R026173/1]. 
XH has received funding from the 
European Union’s Horizon 2020 research and innovation programme under grant agreement No 956123, and is also supported by the UK EPSRC under project [EP/T026995/1].
FW is funded by the Faculty of Environment, Science and Economy at the University of Exeter. PX is funded by the department of computer science at the University of Liverpool. PX contributed to this work equally and conducted this work while she was visiting the University of Exeter. We would like to thank Haozhe Wang, Anjan Dutta, and the anonymous reviewers for their helpful comments and Linyi Li for sharing the pretrained models with us.

\bibliography{aaai23_cr}

\clearpage

\appendix

 
\section{Appendix}
\subsection{Algorithm pseudocode}\label{app:alg}

\begin{algorithm}[!h]\small
\DontPrintSemicolon
  \KwInput{An input example $x$, the objective function $\ell$, the bound of the parameter space $P$, the number of function evaluation $Q$, the number of iterations $T$, the maximum depth $D$,  the size of candidates set $\alpha$}
  \KwOutput{$\ell_{\min}$ with the corresponding solution $c_{\min}$ and an estimation of the ground truth minimum $\ell^*_{\min}$}
  Normalise the parameter space to a unit hypercube with centre point $c_0$\;
  $t \gets 0$, $q \gets 0$\;
  Initialise the index set of hyperrectangles $\mathcal{H} = \{0\}$\;
  Initialise the set of potential optimal space $\mathcal{P}=\{0\}$\;
  \While{$(t \leq T) \cap (q < Q) \cap (\mathcal{P} \neq \varnothing)$}
  {
    Initialise $\mathcal{X} = \{\}$\;
    \For{each potential optimal hyperrectangle $p$ in $\mathcal{P}$}
    {
       \If{hyperrectangle size $\sigma_p = 3^{-D}$}{Continue\;}
       \Else{
          \For{each dimension $i$ with long edge of hyperrectangle $p$}
          {
            $\mbox{Append}(\mathcal{X}, c_p \pm \delta_j^{c_p}\mathbf{e}_i)$\;
            $\mbox{Append}(\mathcal{H}, \{q+1,q+2\})$\; 
            $q \pluseq 2$
          }  
        }
    }
    \tcc{Conduct function evaluation via a single forward propagation}
    $\mathcal{Y} = \ell(\mathcal{X})$\;
    \tcc{Space division}
    \For{each potential optimal hyperrectangle $p$ in $\mathcal{P}$}
    {
        Subdivide hyperrectangle $p$ based on query results in $\mathcal{Y}$\;
        Recording the size $\sigma$ and local slope $\hat{K}$ for all new generated subspaces\;
        Update $p$'s size $\sigma_p$ and local slope $\hat{K}_{c_p}$\;
    }
    \tcc{Record current best evaluation and corresponding solution}
    $\ell_{\min} = \min_{q \in \mathcal{H}}\ell(c_q),\;\mbox{and}\; c_{\min} = \arg\min_{c_q}\ell(c_q)$\;
    Estimate the ground truth $\ell^*_{\min}$ via Eq.~\eqref{eqn:est_minimum}\;
    \tcc{Select potential optimal subspaces}
    Reset $\mathcal{P} = \{\}$\;
    \For{$d \in \{0,1,\ldots,D-1\}$}
    { 
      Build candidates set $\mathcal{H}_\alpha$ from hyperrectangles with $\sigma = 1/3^d$\;
      \For{each hyperrectangle $q$ in $\mathcal{H}_\alpha$}
      {
        \If{$q$ satisfies condition~\eqref{eqn:po_cond2_}}
        {
          $\mbox{Append}(\mathcal{P}, q)$\; 
        }
      }
    }

   	t = t+1	\;
   }
\caption{GeoRobust}\label{alg:workflow}
\end{algorithm}

\subsection{Proofs}

\subsubsection{Proof of Lemma~\ref{lem:lip_stn}\\}

\begin{manualtheorem}{1}
Given an input image example $x \in \mathbb{R}^{H\times W\times C}$ and the ranges of transformation factors, the first-order derivative of geometric transformation with bilinear sampling w.r.t. each transformation factor is bounded.
\end{manualtheorem}

\begin{proof}
The derivative of pixel value $V_i$ w.r.t. $\mathtt{x}_i$ is given by
\begin{equation}\label{eqn:cdvdx_1}\small
\begin{aligned}
\frac{\partial V_i}{\partial \mathtt{x}_i} =&
\sum_{n}^{H} \sum_{m}^W U_{nm}\max(0,1-|\mathtt{y}_i-n|)\\
&\times \begin{cases}
		0  & \mbox{if}\;\; |m - \mathtt{x}_i| \ge 1, \\
		1  & \mbox{if}\;\; |m - \mathtt{x}_i| < 1 \;\;\text{and}\;\; m \ge \mathtt{x}_i,\\
		-1  & \mbox{if}\;\; |m - \mathtt{x}_i| < 1 \;\;\text{and}\;\; m < \mathtt{x}_i.\\
\end{cases}
\end{aligned}
\end{equation}
There are only four neighbouring pixels satisfy $|m - \mathtt{x}_i| < 1$ and $|\mathtt{y}_i - n| < 1$, so Eq.~\eqref{eqn:cdvdx_1} can then be written as
\begin{equation}\label{eqn:cdvdx_2}\small
\begin{aligned}
\frac{\partial V_{i}}{\partial \mathtt{x}_{i}}&= U_{\ubar{n} \bar{m}} \cdot\left(1-y_{i}+\ubar{n}\right) +U_{\bar{n}\bar{m}}  \cdot\left(1-\bar{n}+y_{i}\right)\\
&\quad-U_{\bar{n}\ubar{m}} \cdot \left(1-\bar{n}+y_{i}\right) -U_{\ubar{n} \ubar{m}} \cdot \left(1-y_{i}+\ubar{n}\right)\\
&= \left(1-y_{i}+\ubar{n}\right)\left(U_{\ubar{n} \bar{m}}-U_{\ubar{n} \ubar{m}}\right)\\
&\quad+ \left(1+y_{i}-\bar{n}\right)\left(U_{\bar{n} \bar{m}}-U_{\bar{n} \ubar{m}}\right),
\end{aligned}
\end{equation}
where $(\bar{n},\bar{m}) = (\lceil y_i\rceil, \lceil x_i \rceil)$ and $(\ubar{n},\ubar{m}) = (\lfloor y_i\rfloor, \lfloor x_i \rfloor)$.
We can see that 
\begin{equation}
    \left(1-y_{i}+\ubar{n}\right) + \left(1+y_{i}-\bar{n}\right) = 1, 
\end{equation}
which means Eq.~\eqref{eqn:cdvdx_2} is taking a weighted average of the difference between two pairs of pixels.
Without loss of generality, suppose the eligible pixel value is in $\left[0,1\right]$. We have
\begin{equation}\label{eqn:bouned_dvdx}
    \sup_{\mathtt{x} \in x} (\frac{\partial V_{i}}{\partial \mathtt{x}}) = 1,
\end{equation}
and following a similar deduction, the same result can be obtained for $\frac{\partial V_{i}}{\partial \mathtt{y}}$.
Then, the derivatives of $\mathtt{x}$ and $\mathtt{y}$ w.r.t. transformation matrix $A_\theta$ are
\begin{equation}\label{eqn:dxdA}
    \frac{\partial \mathtt{x}_{i}}{\partial A_\theta} = \left[\begin{array}{ccc}\frac{\partial \mathtt{x}_{i}}{\partial\theta_{11}} \;\;& \frac{\partial\mathtt{x}_{i}}{\partial\theta_{12}} \;\;& \frac{\partial\mathtt{x}_{i}}{\partial\theta_{13}} \\
    0 \;\;& 0 \;\;& 0\end{array}\right] = \left[\begin{array}{ccc}\mathtt{x}_i^\prime \;\;& \mathtt{y}_i^\prime \;\;& 1 \\
    0 \;\;& 0 \;\;& 0\end{array}\right],
\end{equation}
and
\begin{equation}\label{eqn:dydA}
    \frac{\partial \mathtt{y}_{i}}{\partial A_\theta} = \left[\begin{array}{ccc}0 \;\;& 0 \;\;& 0\\
    \frac{\partial \mathtt{y}_{i}}{\partial\theta_{21}} \;\;& \frac{\partial\mathtt{y}_{i}}{\partial\theta_{22}} \;\;& \frac{\partial\mathtt{y}_{i}}{\partial\theta_{23}} 
    \end{array}\right] = \left[\begin{array}{ccc}0 \;\;& 0 \;\;& 0\\
     \mathtt{x}_i^\prime \;\;& \mathtt{y}_i^\prime \;\;& 1\end{array}\right].
\end{equation}
 
For each $\theta$, we have
\begin{equation}\label{eqn:bouned_dxydtheta}
    \sup_{\theta \in A_\theta} (\frac{\partial \mathtt{x}_{i}}{\partial \theta}) = W, 
    \;\mbox{and}\; \sup_{\theta \in A_\theta} (\frac{\partial \mathtt{y}_{i}}{\partial \theta}) = H.
\end{equation}
In the final step, let us take the scaling factor $\lambda$ as an example.
Following the chain rule, the partial derivative is given by
\begin{equation}\label{eqn:dvdlbd}
\frac{\partial V_{i}}{\partial \lambda} =
\frac{\partial V_{i}}{\partial x_{i}} \frac{\partial x_{i}}{\partial \theta_{11}} \frac{\partial \theta_{11}}{\partial \lambda}
+\frac{\partial V_{i}}{\partial y_{i}} \frac{\partial y_{i}}{\partial \theta_{22}} \frac{\partial \theta_{2 2}}{\partial \lambda}.
\end{equation}
Let $\mathcal{R}$ be the set of all eligible $\gamma$, we can substitute Eq.~\eqref{eqn:bouned_dvdx} and~\eqref{eqn:bouned_dxydtheta} into Eq.~\eqref{eqn:dvdlbd} and bound the derivative as
\begin{equation}
    \frac{\partial V_{i}}{\partial \lambda} \leq  \sup _{\gamma \in \mathcal{R}}(\cos \gamma)\cdot(W+H).
\end{equation}
Because there are finite numbers of pixels, the overall derivative has an upper bound as well.
Similarly, by specifying the range of each transformation factor, their derivatives are upper bound correspondingly, and this completes the proof.
\end{proof}

\subsubsection{Proof of Lemma~\ref{lem:find_po}\\}

\begin{manualtheorem}{2}[\cite{Gablonsky01}]
Given the index set $\mathcal{H}$ and a positive tolerance $\tau > 0$. Let $\ell_{\min}$ denote the current best query result. 
Let $\mathcal{H}^p_1 = \{q\in \mathcal{H}:\sigma_q < \sigma_p\}$, $\mathcal{H}^p_2 = \{q\in \mathcal{H}:\sigma_q > \sigma_p\}$ and $\mathcal{H}^p_3 = \{q\in \mathcal{H}:\sigma_q = \sigma_p\}$.
A hyperrectangle $H_p$ is said to be potentially optimal if 
\begin{equation*}
    \ell(c_p) \leq \ell(c_q), \forall q \in \mathcal{H}^p_3,\tag{9}
\end{equation*}
and there is a $\tilde{K} > 0$ such that
\begin{equation*}
\max _{q \in \mathcal{H}^p_{1}} \frac{\ell\left(c_{p}\right)-\ell\left(c_{q}\right)}{\sigma_{p}-\sigma_{q}} \leq \tilde{K} \leq \min _{q \in \mathcal{H}^p_{2}} \frac{\ell\left(c_{q}\right)-\ell\left(c_{p}\right)}{\sigma_{q}-\sigma_{p}},\tag{10}
\end{equation*}
and 
\begin{equation*}
 \begin{cases}
    \tau \leq \frac{\ell_{\min }-\ell\left(c_{p}\right)}{\left|\ell_{\min }\right|}+\frac{\sigma_{p}}{\left|\ell_{\min }\right|} \min _{q \in \mathcal{H}^p_{2}} \frac{\ell\left(c_{q}\right)-\ell\left(c_{p}\right)}{\sigma_{q}-\sigma_{p}},   &  \text{if $\ell_{\min } \neq 0$,} \\
    \ell\left(c_{p}\right) \leq \sigma_{p} \min _{q \in \mathcal{H}^p_{2}} \frac{\ell\left(c_{q}\right)-\ell\left(c_{p}\right)}{\sigma_{q}-\sigma_{p}},
        &  \mbox{otherwise.}
 \end{cases}   \tag{11}             
 \end{equation*}
\end{manualtheorem}

\begin{proof}
For a hyperrectangle $p$, we can group all hyperrectangles into $\mathcal{H}^p_1$, $\mathcal{H}^p_2$, and $\mathcal{H}^p_3$, then inequation~\eqref{eqn:po_cond1} can be rewritten into three inequalities,
\begin{equation}\label{eqn:po_cond1_h1}
\tilde{K} \geq \frac{\ell\left(c_{j}\right)-\ell\left(c_{i}\right)}{\sigma_{j}-\sigma_{i}}, \forall i \in \mathcal{H}^j_{1},
\end{equation}
\begin{equation}\label{eqn:po_cond1_h2}
\tilde{K} \leq \frac{\ell\left(c_{i}\right)-\ell\left(c_{j}\right)}{\sigma_{i}-\sigma_{j}}, \forall i \in \mathcal{H}^j_{2},
\end{equation}
and inequation~\eqref{eqn:po_cond1_h3}.  
Putting inequalities~\eqref{eqn:po_cond1_h1} and~\eqref{eqn:po_cond1_h2} together gives inequity~\eqref{eqn:po_cond1_without_lip}.
If a hyperrectangle satisfies inequalities~\eqref{eqn:po_cond1_h3} and ~\eqref{eqn:po_cond1_without_lip} at the same time, then the PO condition~\eqref{eqn:po_cond1} is satisfied.
While we do not know the true $\tilde{K}$ in Eq.~\eqref{eqn:po_cond2}, it can be replaced by an upper bound given in~\eqref{eqn:po_cond1_h2}.
Substituting condition~\eqref{eqn:po_cond1_h2} into condition~\eqref{eqn:po_cond2} gives us inequalities~\eqref{eqn:po_cond2_}.
This completes the proof.
\end{proof}

\subsubsection{Explanation of Definition~\ref{def:alpha_set}\\}

\begin{figure}[!ht]
    \centering
    \includegraphics[width=0.8\linewidth]{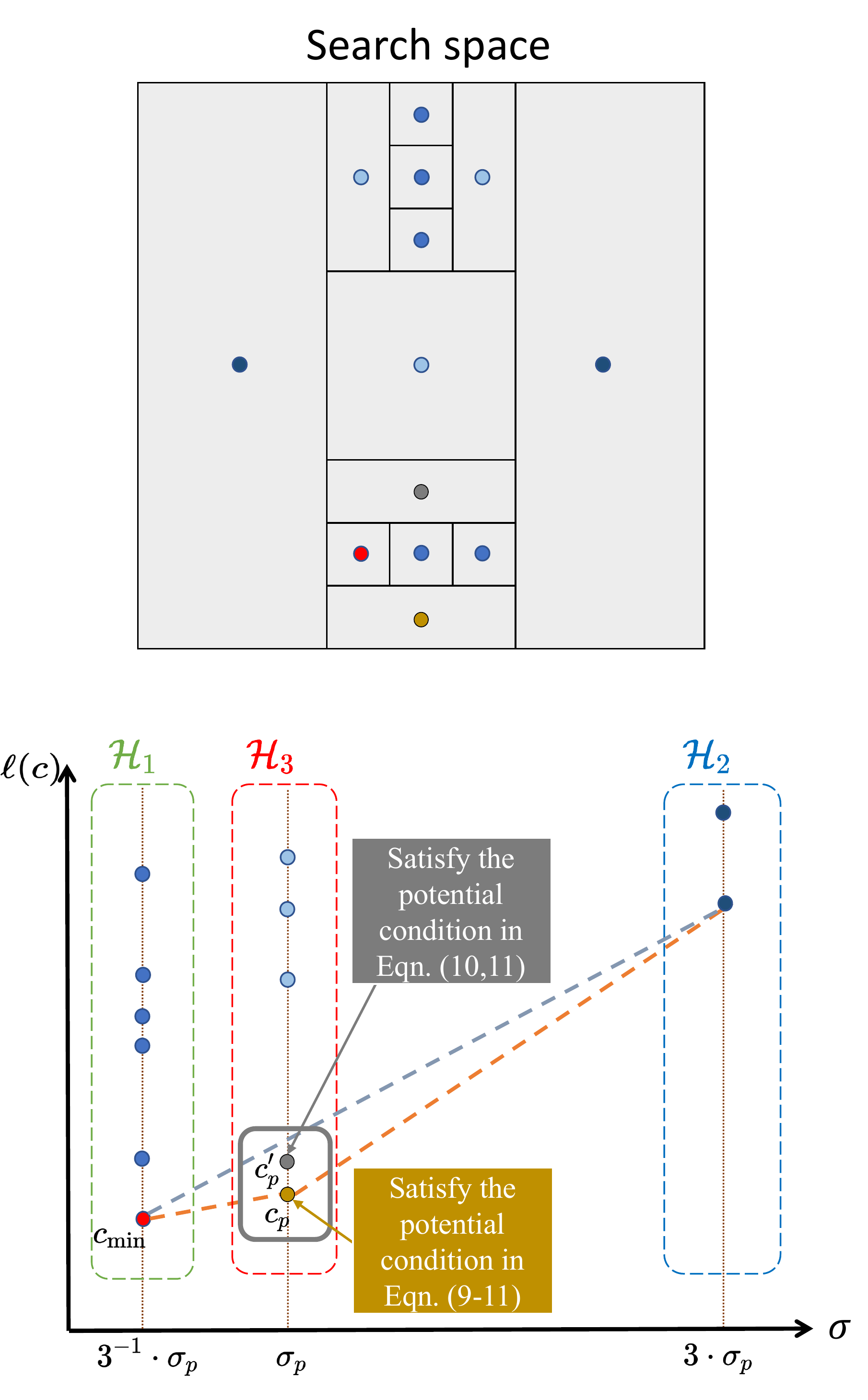}
    \caption{
    A visualisation about potential optimal condition~\eqref{eqn:po_cond1} and  $\alpha$ candidate set ($\alpha=2$).
    A partition of the search space is presented in the upper figure, and the relationship between the sizes and corresponding function values of all subspaces is plotted in the lower figure.
    GeoRobust would select both $c_p$ and $c^\prime_p$ as PO subspaces. 
    }    \label{fig:alpha_set}
\end{figure}

We encourage GeoRobust to select more PO subspace via remove inequation~\eqref{eqn:po_cond1_h3}.
A visualisation of $\alpha$ candidate set is presented in Fig.~\ref{fig:alpha_set}, where both $c_p$ and $c^\prime_p$ would be selected and queried by GeoRobust, while DIRECT optimisation would only choose $c_p$.

\subsubsection{Explanation of Remark~\ref{rmk:all_divided}\\}
In every iteration, there is a hyperrectangle $p$ satisfies  $\sigma_p = \max_{q \in \mathcal{H}}\sigma_q$ and $\ell(c_p) = \min_{q \in \mathcal{H}^p_3} \ell(c_q)$.
Please recall that $\mathcal{H}^p_3$ contains the indexes of hyperrectangles with the same size as $p$ and $\mathcal{H}^p_1$ contains the indexes of hyperrectangles that are smaller than $p$, while $\mathcal{H}^p_2$ is empty because no hyperrectangle larger than $p$ exists.
Because $\mathcal{H}_2^p=\varnothing$, PO condition~\eqref{eqn:po_cond1_without_lip} only produces a lower bound on $\tilde{K}$,~\ie,
\begin{equation}
\max _{i \in \mathcal{H}^j_{1}} \frac{\ell\left(c_{j}\right)-\ell\left(c_{i}\right)}{\sigma_{j}-\sigma_{i}} \leq \tilde{K},
\end{equation}
which means we can always find a slope that is large enough to satisfy PO conditions, making hyperrectangle $p$ a PO subspace.
Therefore, GeoRobust would identify and partition at least one PO hyperrectangle throughout each iteration.
Furthermore, for any hyperrectangle $q$, there is only a finite number of hyperrectangles in its $\mathcal{H}^q_2$ and $\mathcal{H}^q_3$.
Under the worst situation, hyperrectangle $q$ will be selected as a PO space and get divided in the next iteration when $\mathcal{H}^q_2 \cup \mathcal{H}^i_3 = \{q\}$.

\subsubsection{Proof of Theorem~\ref{thm:convergence_complexity}\\}
To prove Theorem~\ref{thm:convergence_complexity}, we need a relationship between the depth of the largest subspace and the number of queries, which is given in the Theorem~4.2 from~\cite{Gablonsky01}.

\begin{manualtheorem1}{4.2}\cite{Gablonsky01}\label{thm:query_depth}
    Assuming that only one hyperrectangle gets divided in every iteration, the number of iterations $T$ after which no hyperrectangle of depth $d-1$ is left is given by
    \begin{equation}\label{eqn:T-with-d}
        T = 3^{n-1}\big(\frac{3^{nd}-1}{3^{n}-1}\big)< 3^{nd}-1.
    \end{equation}
\end{manualtheorem1}
We can now prove Theorem~\ref{thm:convergence_complexity}.
\begin{manualtheorem1}{1}
Let $C$ be the $n$-dimensional united search space and $\tilde{K}$ be the Lipschitz constant of $\ell$ w.r.t. $C$.
The gap between current minima and global minima after $T$ iterations can be written as
\begin{equation*}\small
    \ell_{\min} - \min _{c \in C} \ell(c) \leq \varepsilon < \tilde{K}\cdot(T+1)^{-\frac{1}{n}}.\tag{14}
\end{equation*}
Therefore, to achieve any desired $\varepsilon$, we need up to $\mathcal{O}\big((\tilde{K}/\varepsilon)^n\big)$ iterations.
\end{manualtheorem1}

\begin{proof}
As the global minima must be contained in one of the subspaces, and the objective function is Lipschitz continuous in the search space, we have
\begin{align}\label{eqn:complexity-proof-1}
    \ell_{\min} - \min _{c \in C} \ell(c) &\leq \forall q \in \mathcal{H}, \ell(c_q) - \min _{c \in C} \ell(c)  \\
    &\leq \varepsilon \leq \tilde{K} \cdot 3^{-d},
\end{align} 
where $d$ is the depth of the current largest subspace in the unit search space.
According to Eq.~\eqref{eqn:T-with-d}, we have $d \geq \frac{log_3(T+1)}{n}$, and substituting it into Eq.~\eqref{eqn:complexity-proof-1} gives
\begin{equation}
    \varepsilon \leq \tilde{K}\cdot3^{-\frac{log_3(T+1)}{n}}=\tilde{K}\cdot(T+1)^{-\frac{1}{n}},
\end{equation}
which leads to Eq.~\eqref{eqn:thm-1}.
The relationship between any desired $\varepsilon$ and the number of iterations $T$ is then given by
\begin{equation}
    T \leq (\tilde{K}/\varepsilon)^n - 1.
\end{equation}
We can see that the number of iterations is bound by $\mathcal{O}\big((\tilde{K}/\varepsilon)^n\big)$.
This completes the proof.
\end{proof}

\begin{table*}[t]\small
  \centering
  \caption{Comparison of methods for finding the worst-case transformation and providing the lower bound for verification.} \label{tab:related_work}%
  \scalebox{0.8}{
    \begin{tabular}{lccccccc}
    \toprule
    \multicolumn{1}{c}{\multirow{2}[1]{*}{Method}} & \multicolumn{1}{c}{\multirow{2}[1]{*}{Approach }} & \multicolumn{1}{c}{\multirow{2}[1]{*}{Requirement}} & \multicolumn{1}{c}{\multirow{2}[1]{*}{Efficiency}} & \multicolumn{2}{c}{Scalability} & \multicolumn{2}{c}{Guarantee} \\
    \cmidrule(r){5-6}\cmidrule(r){7-8}
    \multicolumn{1}{c}{} &       & &      & \multicolumn{1}{c}{Architecture} & \multicolumn{1}{c}{Scale} & \multicolumn{1}{c}{Lower bound} & \multicolumn{1}{c}{Worst-case} \\
    \midrule
    Exhaustive search~\cite{PeiCYJ17} & Query & None  & \XSolid   & \Checkmark & \Checkmark & \XSolid  & \Checkmark \\
    Random pick~\cite{EngstromTTSM19} & Query & None  & \Checkmark & \Checkmark   &  \Checkmark  & \XSolid & \XSolid \\
    DeepG~\cite{BalunovicBSGV19} & \makecell[c]{Layer-by-layer\\ propagation} & \makecell[c]{Specify transformation \\access all parameters} &   \XSolid  &  \XSolid   &  \XSolid &  \Checkmark  & \XSolid \\
    Semantify-NN~\cite{MohapatraWCLD20} & \makecell[c]{Surrogate network\\and layer-by-layer\\ propagation} & \makecell[c]{Specify transformation \\access all parameters} & \XSolid  &  \XSolid   &  \XSolid &  \Checkmark  & \XSolid \\
    DistSPT~\cite{FischerBV20} & \makecell[c]{Random smoothing\\and Layer-by-layer\\ propagation} & \makecell[c]{Specify transformation \\access all parameters} & \XSolid  &  \Checkmark   &  \Checkmark &  \Checkmark  & \XSolid \\
    TSS~\cite{LiWXRK00021} & \makecell[c]{Random smoothing} & \makecell[c]{Specify transformation} & \XSolid  &  \Checkmark   &  \Checkmark &  \Checkmark  & \XSolid \\
    GSmooth~\cite{HaoYD0SZ22} & \makecell[c]{Surrogate network\\ and random smoothing} & \makecell[c]{Specify transformation \\access all parameters} & \XSolid  &  \Checkmark   &  \XSolid &  \Checkmark  & \XSolid \\
    \midrule
    GeoRobust (ours) & Query & None  &    \Checkmark  &  \Checkmark    &    \Checkmark  &  \Checkmark &  \Checkmark\\
    \midrule
    \end{tabular}%
    }
\end{table*}%

\subsection{Detailed related works}

Numerous studies have been conducted to find the worst-case adversarial perturbation.
While several adversarial attacks, such as the projected gradient descent attack~\cite{MadryMSTV18}, and Auto Attack~\cite{croce2020reliable}, can generate strong adversarial examples, they cannot ensure finding the worst-case perturbation~\cite{HuangKR+18}.
Some complete verification technologies can be used to find the worst-case perturbation~\cite{liu2021algorithms}, where completeness means that a method is guaranteed to find adversarial examples within a given norm ball unless no adversarial example exists, but most of them are computationally inefficient and have specific requirements for their target models.
ExactReach~\cite{xiang2017reachable} and ReluVal~\cite{wang2018formal}, for example,  perform layer-by-layer propagation through target models with only linear or ReLU activations, requiring their target models to be fully accessible.
Therefore, these methods only work under the white-box setting and are unsuitable for large-scale neural networks.
Apart from the limitation on scalability, the layer-by-layer propagation operation needs a $L_p$ norm based pixel-level or element-level bounding box of the input.
As illustrated in Fig.~\ref{alg:workflow}, it is difficult to establish such a bounding box for geometric transformations because even a small transformation could affect a huge number of pixels and drastically alter their value.
DeepGo~\cite{RuanHK18} is a global optimisation based method that operates under the grey-box environment,~\ie, requiring no knowledge of the model's parameters but a pre-estimation of the model's Lipschitz constant, which is difficult to get in reality.
Due to space limitations, we cannot cover all complete verification methods here and refer readers to a recent survey on verification techniques~\cite{liu2021algorithms}.

On the other hand, there are also some studies on the geometric robustness of DNNs, and we summarise the difference between our method and related works in Table~\ref{tab:related_work} on finding the worst-case geometric transformation.
\citet{JaderbergSZK15} proposed a differential module called spatial transformer network (STN) to enhance neural networks' learning ability regarding geometric transformations. 
Although the parameter space of control factors for most geometric manipulations is continued, the image pixels' coordinates are discrete, bounded integers, which means the possible outcomes for a particular set of transformations are finite. 
Pei~\etal~\cite{PeiCYJ17} empirically evaluated DNNs' resistance toward geometric transformations by enumerating all possible values.
Similarly, Engstrom~\etal~\cite{EngstromTTSM19} employed random pick and grid search to discover the adversarial translation and rotation to deceive target models.
DeepG~\cite{BalunovicBSGV19} computes a convex relaxation of the bounding box for a set of geometric transformations and then certifies the robustness property via existing robustness verifier~\cite{SinghGPV19}.
Mophapatra~\etal~\cite{MohapatraWCLD20} introduced a small network, called Semantify-NN, to simulate the geometric manipulation and adopted existing verifier~\cite{weng2018FastLin} to examine the hybrid model composed of Semantify-NN and a target network. 
Because FastLin~\cite{weng2018FastLin} and DeepPoly~\cite{SinghGPV19} are incompleteness verifiers, these two works may certify whether a set of geometric transformations can affect the predictions of a target classifier but are unable to determine the worst-case transformations precisely.
Besides, these two verifiers use layer-by-layer propagation, which is computationally inefficient and limited to small networks in the white-box setting.
DistSPT~\cite{FischerBV20}, TSS~\cite{LiWXRK00021}, and GSmooth~\cite{HaoYD0SZ22} utilise random smoothing techniques to verify the geometric robustness.
TSS is a black-box verification method that is based on random smoothing.
DistSPT combines random smoothing and interval bound propagation together to conduct the verification on tasks beyond $L_p$ norm.
GSmooth also uses an image-to-image network to simulate the geometric transformation.

Parallel to geometric transformation, several works~\cite{AlaifariAG19,XiaoZ0HLS18} investigated spatial transformation, which is a spatial distortion of the coordinates of pixels.
Please note that spatial transformations performed using vector fields remain pixel-level perturbation. 
Thus, it is fundamentally distinct from geometric transformation and beyond the scope of this paper.
\begin{table}[!ht]\small
  \centering
  \caption{Benign accuracy of models trained and verified by~\citet{LiWXRK00021}. The small CNN used for MNIST classification has 4 convolutional layers and 3 fully connected layers.}\label{tab:app_benign_tss_acc}%
  \scalebox{0.9}{
    \begin{tabular}{ccccc}
    \toprule
    Model & Dataset & Transf. & Reported acc. & Reproduced acc. \\
    \midrule
    \multirow{2}[2]{*}{small CNN} & \multirow{2}[2]{*}{MNIST} &  $R(50^{\circ})$   &  99.4\%   & 99.4\% \\
          &       & $S(0.3)$  & 99.4\%      & 99.4\% \\
    \midrule
    \multirow{3}[2]{*}{ResNet101} & \multirow{3}[2]{*}{CIFAR-10} & $R(10^{\circ})$ &    83.2\%   & 84\% \\
          &       & $R(30^{\circ})$      &  82.6\%     & 81.2\% \\
          &       & $S(0.3)$   &   79.8\%    & 80.8\% \\
    \midrule
    \multirow{2}[2]{*}{ResNet50} & \multirow{2}[2]{*}{ImageNet} &$R(30^{\circ})$   &  46.2\%     & 20.8\% \\
          &       & $S(0.3)$   &  50.8\%     & 26.6\%  \\
    \bottomrule
    \end{tabular}
    }%
\end{table}%

\subsection{Experiments}

\begin{table*}[!ht]\small
  \centering
  \caption{
  Verifying geometric robustness on ImageNet against all combination of three transformations:$R(20^{\circ})$, $S(0.1)$, and $T(22.4,22.4)$. The target model is ResNet50, which achieves 74\% classification accuracy.
  To make a fair comparison on efficiency, the random pick and grid search are also implemented on GPU, where the batch size is 128.
  }\label{tab:app_all_transf}
  \scalebox{0.9}{
    \begin{tabular}{cccccccccc}
    \toprule
    \multirow{2}[2]{*}{Transformations} & \multicolumn{3}{c}{GeoRobust} & \multicolumn{3}{c}{Random pick} & \multicolumn{3}{c}{Grid search} \\
    \cmidrule(r){2-4}\cmidrule(r){5-7}\cmidrule(r){8-10}
          & Verified Acc. & \#Queries & Runtime (s) & Verified Acc. & \#Queries & Runtime (s) & Verified Acc. & \#Queries & Runtime (s) \\
    \midrule
    $R$     &   58\%    &   667 $\pm$ 29    &    2.4    &   58\%    &   \multirow{2}[2]{*}{2000}    & 4.8      &  58\%      & \multirow{2}[2]{*}{2000}      & 4.7  \\
    $S$     &    59\%   &   679 $\pm$ 30    &   2.4     &   59\%    &      & 4.7       &    59\%    &       & 4.7 \\
    \midrule
    $R+S$   &    \textbf{54\% }  &   1096 $\pm$ 151    &   3.8    &   56\%    & \multirow{2}[2]{*}{4000}      &   9.6     &   56\%    & \multirow{2}[2]{*}{$5000$}      &  12.1 \\
    $T$     &    57\%   &   1046 $\pm$ 99     &   3.6    &    57\%    &       &   9.4     &  59\%     &       &  11.9 \\
    \midrule
    $R+T$   &     \textbf{46\%}  &  1187 $\pm$ 112     &   4.1     &   51\%   &  \multirow{2}[2]{*}{6000}     & 14.4      &      49\% &\multirow{2}[2]{*}{$1\times10^4$}       &  29.3 \\
    $S+T$   &   \textbf{57\% }   &   1170 $\pm$ 111    &   4.0     &   60\%     &       & 14.1       &   57\%    &       & 28.6 \\
    \midrule
    $R+S+T$ &    \textbf{46\%}   &    1295 $\pm$ 117   &   4.5     &   49\%    &   8000    &   19.1    &   46\%    &   $1\times10^5$    & 251.3 \\
    \bottomrule
    \end{tabular}}%
\end{table*}%

\subsubsection{Implementation details within Tab.~\ref{tab:baseline}\\}

To make the comparison, we use GeoRobust to verify the same models used in~\cite{LiWXRK00021} on the same subsets of MNIST, CIFAR-10, and ImageNet.
In Tab.~\ref{tab:app_benign_tss_acc}, we present the benign accuracy reported by~\cite{LiWXRK00021} and obtained on our machine.
It can be seen that the reproduced accuracy of MNIST and CIFAR-10 models are basically consistent with the reported accuracy, while the ImageNet models are much less accurate than expected (corresponding code is provided for reviewing).
Since we failed to load the ImageNet models properly, the comparison was only done on MNIST and CIFAR-10 datasets in Tab.~\ref{tab:baseline}.

GeoRobust is carried out with $D=5$ and $\alpha=2$, and its computational budget is up to 200 iterations and 2000 queries.
In practice, GeoRobust only conducted 244 queries on average to verify the 1-dimensional adversarial geometric transformations. 
The average runtime on MNIST and CIFAR-10 are 0.18 and 0.74 seconds, respectively.
The grid search, here, is carried out with 2000 queries, which is sufficient for exploring 1-dimension transformation.

\paragraph{Additional experiments on all combinations of geometric transformation}
In Tab.~\ref{tab:imagenet_baseline}, we compared GeoRobust to random pick and grid search under four combinations of geometric transformations, while the comparison on all combinations of geometric transformations is summarised in Tab.~\ref{tab:app_all_transf}.
GeoRobust is carried out with $D=6$ and $\alpha=2$, and the computational budget is up to 150 iterations and 2000 queries.
It can be seen from Tab.~\ref{tab:app_all_transf} that GeoRobust is significantly more efficient than random pick and grid search under all settings.
While three methods perform similarly on verifying 1-dimensional geometric transformations, GeoRobust can achieve the same and sometimes even better performance than grid search when verifying multiple transformations together.

\paragraph{Detailed benchmark on ImageNet classifiers}
We present a completed version of Tab.~\ref{tab:benchmark_imagenet}, presenting the average numbers of queries and runtime.
Here GeoRobust can run up to 150 iterations and 3000 queries per example.
The depth and candidate set are set to be $D=6$ and $\alpha=2$.
The geometric transformations are $R(20^{\circ})$, $S(0.1)$, and $T(22.4,22.4)$.
In Tab.~\ref{tab:app_imagenet}, we can see that GeoRobust can conduct super efficient analysis on all ImageNet classifiers, and it only takes GeoRobust less than 11 seconds to analyse one example on the large Vit with $3\times 10^8$ parameters, which is the largest model here.

\begin{table*}[!ht]
\centering
\caption{A completed version of Tab.~\ref{tab:benchmark_imagenet}: Benchmarking Geometric Robustness on ImageNet}\label{tab:app_imagenet}
\scalebox{0.9}{
\begin{tabular}{lcccccc}
\toprule[0.8pt]
Models & Clean& Attack & Verified& \#Parameters & \# Queries& Runtime (s)\\
\midrule

Inception v3.                    &73.60\%&28.20\%&24.20\%&2.4$\times10^7$&1405$\pm$205&4.6$\pm$0.5\\
Inception v$3_{adv}$             &75.00\%&30.60\%&27.00\%&2.4$\times10^7$&1398$\pm$192&4.6$\pm$0.5\\
Inception v4                     &78.40\%&40.20\%&36.40\%&4.3$\times10^7$&1444$\pm$256&6.3$\pm$0.9\\
\midrule             
ResNet34                         &64.40\%&10.60\%&9.00\%&2.2$\times10^7$&1475$\pm$245&3.1$\pm$0.4\\
ResNet50                         &78.40\%&54.00\%&31.12\%&2.6$\times10^7$&805$\pm$110&4.8$\pm$0.6\\
Wide ResNet50.                   &81.60\%&49.40\%&40.00\%&6.9$\times10^7$&1283$\pm$125&6.0$\pm$0.5\\
ResNet101                        &80.00\%&54.20\%&48.20\%&4.5$\times10^7$&1291$\pm$134&5.8$\pm$0.5\\
ResNet152                        &79.40\%&53.80\%&46.20\%&6.0$\times10^7$&1278$\pm$129&7.3$\pm$0.6\\

\midrule
$\mbox{Vit}_{32\times32}$        &75.60\%&23.40\%&19.00\%&8.8$\times10^7$&1528$\pm$297&3.3$\pm$0.5\\
$\mbox{Vit}_{16\times16}$        &81.40\%&41.20\%&34.20\%&8.6$\times10^7$&1471$\pm$268&5.0$\pm$0.8\\
Large $\mbox{Vit}_{16\times16}$  &83.40\%&49.20\%&40.20\%&3.0$\times10^8$&1410$\pm$244&10.4$\pm$1.6\\
\midrule
$\mbox{Beit}_{16\times16}$.      &83.80\%&56.40\%&52.00\%&6.5$\times10^7$&1403$\pm$215&4.9$\pm$0.7\\
Large $\mbox{Beit}_{16\times16}$ &\textbf{85.60\%}&\textbf{65.60\%}&\textbf{58.20\%}&2.3$\times10^8$&1363$\pm$190&10.5$\pm$1.3\\
\midrule
Gmlp                             &77.96\%&40.80\%&36.80\%&1.9$\times10^7$&1661$\pm$417&4.3$\pm$0.9\\
Mixer                            &72.20\%&27.20\%&23.40\%&6.0$\times10^7$&1566$\pm$337&4.5$\pm$0.9\\
Swin                             &80.20\%&34.60\%&13.20\%&8.8$\times10^7$&1292$\pm$136&5.4$\pm$0.5\\
Xcit                             &76.80\%&40.40\%&20.60\%&8.4$\times10^7$&1497$\pm$355&6.5$\pm$1.1\\
Pit                              &79.40\%&36.60\%&20.00\%&7.4$\times10^7$&1538$\pm$371&11.0$\pm$1.1\\
\bottomrule
\end{tabular}}
\end{table*}

\end{document}